# An End-to-End Comprehensive Gear Fault Diagnosis Method Based on Multi-Scale Feature-Level Fusion Strategy

Bowei Qiao[1]，Hongwei Wang[*]

（Xinjiang University, School of Intelligent Manufacturing and Modern Industry, Urumqi, 830000, China.）

# Abstract

Bowei Qiao，Hongwei Wang [*] ，Zhao Li ，Zhiyun Dou ，Honggang Liu

(College of Intelligent Manufacturing Modern Industry, Xinjiang University, Urumqi 830000, China)

**Abstract:** To satisfy the requirements of the end-to-end fault diagnosis of gears, an integrated intelligent method of fault diagnosis for gears using acceleration signals was proposed, which was based on Gabor-based Adaptive Short-Time Fourier Transform (Gabor-ASTFT) and Dual-Tree Complex Wavelet Transform(DTCWT) algorithms, Dilated Residual structure and feature fusion layer, is proposed in this paper. Initially, the raw one-dimensional acceleration signals collected from the gearbox base using vibration sensors undergo pre-segmentation processing. The Gabor-ASTFT and DTCWT are then applied to convert the original one-dimensional time-domain signals into two-dimensional time-frequency representations, facilitating the preliminary extraction of fault features and obtaining weak feature maps.Subsequently, a dual-channel structure is established using deconvolution and dilated convolution to perform upsampling and downsampling on the feature maps, adjusting their sizes accordingly. A feature fusion layer is then constructed to integrate the dual-channel features, enabling multi-scale analysis of the extracted fault features.Finally, a convolutional neural network (CNN) model incorporating a residual structure is developed to conduct deep feature extraction from the fused feature maps. The extracted features are subsequently fed into a Global Average Pooling(GAP) and a classification function for fault classification. Conducting comparative experiments on different datasets, the proposed method is demonstrated to effectively meet the requirements of end-to-end fault diagnosis for gears.

**Keywords:** Fault Diagnosis of gears; Gabor-based Adaptive Short-Time Fourier Transform(Gabor-ASTFT); Dual-Tree Complex Wavelet Transform(DTCWT); Deep learning; feature fusion;

## 1. Introduction

Here is the translation in the SCI paper format with bracketed citations:

As critical components of rotating machinery, gear bearings are widely used in industrial production and engineering fields [1]. Due to the inevitable occurrence of faults and downtime during the operation of mechanical equipment, which can lead to economic losses and personnel injuries, fault diagnosis is of significant importance for rotating machinery [2-3]. To ensure the reliability and safety of mechanical equipment, it is essential to monitor the operational status of gear bearings. When a fault occurs in one component of rotating machinery, it may affect other operating components. At certain fault frequencies, the acceleration signal may exhibit cyclical impact modulation. Monitoring the gearbox acceleration signal is one of the most common methods [5-6]. In the field of fault diagnosis, traditional fault monitoring methods mainly rely on knowledge-driven approaches, using signal processing algorithms such as STFT, DWT, and EMD to analyze vibration data from rotating machinery for fault diagnosis [7]. These traditional signal analysis methods have achieved some success [8]. However, in most cases, the acceleration signals collected from the gearbox base are non-stationary, containing significant noise and complex modulation components, making it difficult to perform mechanical analysis on the signals and leading to errors in fault diagnosis results. Moreover, frequency-based identification methods heavily rely on expert knowledge, and generating spectra and analyzing frequency contents can be time-consuming [9-10]. With the development of artificial intelligence, modern fault diagnosis methods for gear bearings mainly adopt a data-driven approach [11-12]. Gu, J. used CNN models and SVM for fault classification and achieved high recognition accuracy [13]. Shan, S. applied the Mel-CNN model for bearing fault diagnosis and used the proposed Mel-CNN model on motor noise data containing bearing faults [14]. Chen, Z. introduced a new deep learning fault diagnosis method based on cyclic spectral coherence (CSCoh) and convolutional neural networks (CNN) for two-dimensional representations, aiming to improve the fault identification performance for rolling bearings [15]. These methods have achieved high fault diagnosis accuracy, and CNN models with feature fusion also contribute to bearing fault diagnosis. Zhang, Z. employed a novel bearing intelligent fault diagnosis model using dual-layer data fusion (DLDF), providing a comprehensive representation of bearing fault characteristics [16]. Xu, Y. proposed a new convolutional fusion framework using collaborative fusion convolutional neural networks (CFCNN) to improve fault diagnosis accuracy [17]. Shao, H. used a stacked wavelet autoencoder structure and fusion strategy to provide more accurate and reliable fault diagnosis results from multi-sensor noise vibration signals [18]. Yan, J. used a second-order convolutional neural network (QCNN) for bearing fault diagnosis with audio and vibration signals [19]. Zhang, L. applied a fusion strategy to help improve fault diagnosis accuracy for wind turbine bearings under complex scenarios [20].However, most modern studies focus on fault diagnosis of bearings in rotating machinery, while fault diagnosis of gear signals is relatively less explored. Additionally, the analysis methods for fault signals are often limited to a single approach, lacking multi-scale or multi-method fusion for feature classification. As a result, there is still room for improvement in fault diagnosis accuracy.

Based on the above research,in this paper,an integrated intelligent method of fault diagnosis for gears using acceleration signals was proposed,which was based on Gabor-ASTFT and DTCWT algorithms, Dilated Residual structure and feature fusion layer. By monitoring the acceleration signals of a planetary gearbox, multi-channel data acquisition is employed to collect fault signals from different types of faulty gears under specific load conditions. The Gabor-ASTFT and DTCWT are utilized to extract features from the raw signals. To refine the weak features obtained, dilation convolution and transposed convolution are applied for sampling adjustment. A feature layer is introduced for feature fusion, and a residual CNN model is constructed for fault classification. Finally, comparative experiments are conducted on different datasets to evaluate the proposed method.The contributions of this paper are as follows:

1.An improved CNN model, based on Dilated Residual Networks (DRN), is proposed for the extraction of subtle features from sensor signals. The trained model is capable of extracting features from raw signals, and high classification accuracy is demonstrated across various datasets.

2.A feature fusion strategy is introduced, where features extracted from different convolutional channels are combined in a fusion layer. The incorporation of feature fusion significantly enhances the classification accuracy of the model.

3.The combination of Gabor-based Adaptive Short-Time Fourier Transform (Gabor-ASTFT) and Dual-Tree Complex Wavelet Transform(DTCWT) enhances the ability to extract time-frequency features from complex,non-stationary signals, providing a more comprehensive representation of the signal characteristics.

The sections of this paper are arranged as follows:the basic theories are mainly elaborated in the Sections 2 and 3.The main procedure of the proposed fault diagnosis method of gears is explained in Section 4.Then,the proposed method is validated with different datasets and comparative experiments in Section 5.Finally,the conclusions are drawn in Section 6.

## 2. Signal processing

### 2.1. Gabor-Based Adaptive Short-Time Fourier Transform (Gobar-ASTFT)

The Gabor-based Adaptive Short-Time Fourier Transform (Gabor ASTFT) is used to process one-dimensional fault signal segments, which is an improved method based on the Short-Time Fourier Transform (STFT). In the field of fault diagnosis, STFT is one of the most fundamental time-frequency analysis methods and is widely used. Compared with the traditional Fourier transform, STFT achieves local time-frequency analysis by applying a sliding window to segment the original time-domain signal, which enables the extraction of fault features from the fault signal, carrying richer information and helping to improve the accuracy of fault diagnosis models. Moreover, vibration fault signals are generally non-stationary, while traditional Fourier transforms only provide a global frequency-domain representation, suitable for stationary signals. However, STFT can effectively handle non-stationary signals, such as pulse signals and transient vibration signals, by introducing window functions, making it more suitable for extracting vibration fault signals. The STFT is given by the following formula:

$$STFT(t,\omega) = \int_{\infty}^{\infty} h(u)f(t+u)e^{ju\omega}du \tag{1}$$

Where $t$ is the time, $\omega$ is the frequency，$f(t)$ is the raw signal in the time domain,and $h(u)$ is a window function such as rectangular, Gaussian,Hanning,etc.

However, according to the Heisenberg Uncertainty Principle,

$$\Delta t \cdot \Delta f \geq \frac{1}{4\pi} \tag{2}$$

where $\Delta t$ and $\Delta f$ represent the time and frequency resolution, respectively. The product of time-frequency resolution has a lower limit, and the Gaussian window used in the Gabor transform can precisely reach the lower limit of the uncertainty principle, achieving optimal time-frequency localization. Additionally, rectangular windows and Hanning windows in STFT can cause signal truncation, turning periodic signals into non-periodic ones, introducing additional high-frequency components in the frequency domain, leading to spectral leakage and reducing resolution accuracy. However, the Gaussian function and its Fourier transform are also Gaussian functions, which result in excessive smoothing in the time domain and low sidelobes in the frequency domain, avoiding boundary effects and reducing spectral leakage.

When analyzing gear acceleration signals, the selection of the window function must consider different frequency components:(a)For high-frequency transient components, a narrow window is required to achieve high time-domain resolution.(b)For low-frequency steady-state components, a wide window is needed to provide high frequency-domain resolution.

However, the traditional Gabor transform utilizes a fixed-length window, which limits its adaptability. To overcome this limitation, the ASTFT method is incorporated. By utilizing techniques such as time-frequency entropy (TFE), instantaneous frequency (IF), energy distribution, or optimization algorithms, the Gabor transform window is adaptively adjusted to dynamically optimize its length.

In this study, the Particle Swarm Optimization (PSO) algorithm, inspired by the foraging behavior of bird swarms and based on swarm intelligence, is employed to optimize the adaptive Gabor window.

---

**Algorithm 1：** PSO based on Wigner-Ville Distribution(WVD)

Definition of the Optimal Window and Ideal Distribution(WVD Distribution)

$$P_{aim}(t,f) = \int_{-\infty}^{\infty} x(t+\frac{\tau}{2})x^*(t-\frac{\tau}{2})e^{-j2\pi f\tau}d\tau$$

Definition of Particles, Particle Velocity, and Swarm Parameters.

$$L_i(t) = [L_i(1), L_i(2), \cdots, L_i(T)], L_{min} \leq L_i(t) \leq L_{max}$$

$$V_i(t) = [V_i(1), V_i(1), \cdots, V_i(T)]$$

Calculate fitness.

---

$$F(L) = -\sum_t \sum_f \left| P_{ASTFT}(t,f) - P_{aim}(t,f) \right| \quad F(L) = -\sum_t \sum_f \left| P_{ASTFT}(t,f) - P_{aim}(t,f) \right|$$

Update particle parameters.

$$V_i^{k+1}(t+1) = \omega V_i^k(t) + c_1 r_1 (P_{best,i} - L_i^k(t)) + c_2 r_2 (G_{best} - L_i^k(t))$$

$$L_i^{k+1}(t+1) = L_i^k(t) + V_i^{k+1}(t+1)$$

Iterations $\rightarrow$ Convergence or max-iteration, then get the optimal window size.

$$L^*(t) = G_{best}$$

For each signal, 10 repeated experiments were conducted to take the mode, in order to avoid getting trapped in local optima. The window length range was [16, 127], the particle swarm size was 30, and the maximum number of iterations was 20. A segment of the raw signal was divided into 16 smaller sections, each subjected to the optimal window length search, ultimately achieving convergence. One of the optimal window search processes is shown in Figure 1.

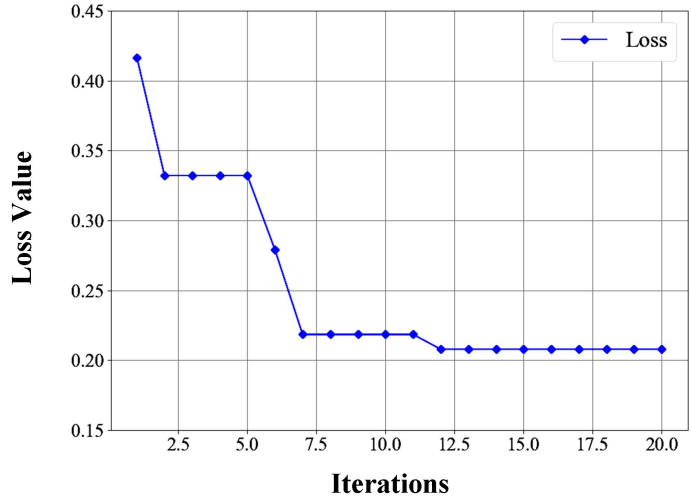

**Figure 1.** Diagram of the PSO Process for Searching the Optimal Window

## 2.2. Dual-Tree Complex Wavelet Transform(DTCWT)

The Dual-Tree Complex Wavelet Transform (DTCWT) is used in this paper to process one-dimensional fault signal segments from other channels, which is an improved method based on wavelet transform. In the field of fault diagnosis, wavelet transform is widely applied. It enables time-frequency analysis of vibration signals. By adjusting the scale and translation parameters of the wavelet, different wavelet bases can be obtained, allowing multi-scale analysis of the original time-series signal at different time-frequency resolutions. However, the Continuous Wavelet Transform (CWT) is a high-complexity redundant transform with poor time resolution in high-frequency regions and poor frequency resolution in low-frequency regions. The Discrete Wavelet Transform (DWT) reduces redundancy but has lower resolution for high-frequency signals. The Wavelet Packet Transform (WPT) overcomes this limitation by decomposing both low-frequency and high-frequency components at each level, thus improving high-frequency resolution. However, due to downsampling and splitting operations, both DWT and WPT inherently exhibit shift variance. This means that a slight shift in the input signal may cause significant changes in the energy distribution of wavelet coefficients at different scales, potentially affecting the effectiveness of detecting transients related to machine faults.

In contrast, the Dual-Tree Complex Wavelet Transform (DTCWT) provides approximate shift invariance, effectively addressing this issue. Additionally, DTCWT exhibits higher time-frequency resolution in both low-frequency and high-frequency regions. It also inherits the advantages of complex wavelet transform by using two orthogonal wavelet trees to separately calculate the real and imaginary parts of the signal, and then combining them into the final complex wavelet coefficients. Compared to real-valued wavelets, DTCWT provides additional phase information, offering a richer perspective for fault detection. The DTCWT is given by the following formula:

$$\begin{cases} W_{s,k} = \sum_{n=-\infty}^{\infty} f(n)\psi_{s,k}(n) \\ W_{dt}(s,k) = W_A(s,k) + jW_B(s,k) \\ M(s,k) = \sqrt{W_A^2(s,k) + W_B^2(s,k)} \\ \theta(s,k) = \tan^{-1}\left(\frac{W_B(s,k)}{W_A(s,k)}\right) \end{cases} \tag{3}$$

where $W_A$ and $W_B$ represent the real and imaginary outputs of the wavelet trees, respectively. $M$ denotes the magnitude, $\theta$ represents the phase, and $s$ and $k$ correspond to the scale and translation parameters, respectively.

## 3. Model construction

### 3.1 Fusion layer

The fused features provide a more comprehensive representation of the machine's condition, thereby enhancing fault detection and identification.Feature fusion is commonly categorized into three approaches: data-level fusion, feature-level fusion, and decision-level fusion. Compared to data-level fusion, feature-level fusion focuses on the weak features extracted from raw signals, reducing the computational burden associated with redundant signal processing. In contrast to decision-level fusion, feature-level fusion is easier to implement and generally achieves superior performance [22].

At the feature-level fusion stage, due to the mismatch in feature map sizes resulting from different feature extraction methods applied to the original vibration signals, direct fusion is infeasible. To address this issue, an upsampling convolutional path and a downsampling convolutional path are designed to adjust the input Gabor-ASTFT feature maps and DTCWT feature maps before fusion, ensuring correct feature alignment for effective integration.

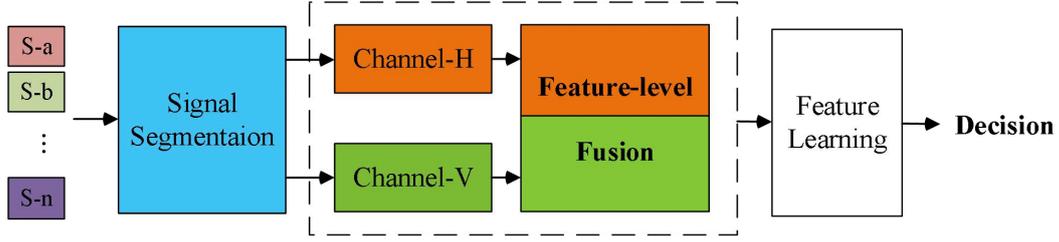

**Figure 2.** Feature-level fusion

Feature-level fusion typically employs various techniques, including feature concatenation, feature selection, principal component analysis (PCA) [23], canonical correlation analysis (CCA) [24], , and deep learning-based fusion methods [25]. In this study, a deep learning-based fusion approach is adopted, where the Gabor-ASTFT and DTCWT methods are used to extract weak features from the raw signals. A deep learning model for feature fusion is then constructed to perform classification.

### *3.2 Dilated convolutions*

Dilated convolutions are a specialized type of convolutional operation that have been widely applied in time-series signal processing, image segmentation, and generative adversarial networks (GANs)[26]. In convolutional neural networks (CNNs), they serve as an augmentation technique that expands the receptive field of the convolutional kernel without increasing computational cost or kernel size by inserting gaps between the kernel elements. In conventional convolution, the kernel performs a dot-product operation with the input feature data. Let the input feature map be $X \in R^{H' \times W'}$, the convolution kernel be $K \in R^{kH \times kW}$, and the output feature map be $Y \in R^{H \times W}$; $x(i,j)$ denotes the input signal, $y(i,j)$ represents the convolution output, $\omega(m,n)$ represents the kernel weights, $K_w$ and $K_h$ denotes the kernel size.

In a dilated convolution layer, the convolution kernel does not densely cover every input element as in traditional convolution. Instead, by introducing a gap defined by the dilation rate, the receptive field of the kernel is effectively enlarged. By increasing the dilation rate, a one-dimensional dilated convolution can capture a broader range of temporal information, while a two-dimensional dilated convolution can encompass larger time–frequency information—thus enabling each convolutional output to incorporate a wider range of features. In this study, we process two-dimensional feature maps using dilated convolutions:

$$\begin{cases} y(i,j) = \sum_{m=0}^{K_h-1} \sum_{n=0}^{K_w-1} x(i+r_1 \cdot m,\ j+r_2 \cdot n) \cdot \omega(m,n) \\ H = \left[ \dfrac{H' - (K_h + (K_h - 1) \cdot (r_1 - 1))}{S} + 1 \right] \\ W = \left[ \dfrac{W' - (K_w + (K_w - 1) \cdot (r_2 - 1))}{S} + 1 \right] \end{cases} \qquad (4)$$

where $r_1$ and $r_2$ denote the dilation factors along the time and frequency dimensions, respectively. All other parameters remain the same as in conventional convolution. It is evident that dilated convolution only alters the computational coverage of the kernel on the input features—thereby expanding the receptive field—without changing the kernel size or increasing the computational load. In our work, dilated convolutions are applied to the input wavelet transform feature maps; this approach increases the feature extraction range of the convolutional layers over the time–frequency feature maps without augmenting network parameters, thus facilitating feature extraction and down-sampling by resizing the convolved feature maps to a suitable dimension for feature fusion.

### 3.3 Deconvolution

Deconvolution, also known as transposed convolution, is a commonly used up-sampling operation in convolutional neural networks and finds widespread application in image generation[27], image resolution restoration, and related fields. Unlike conventional convolution, which typically reduces the spatial dimensions of feature maps after successive layers, deconvolution serves as an inverse operation that projects low-resolution feature maps back into a high-dimensional space, thereby generating higher-resolution output images or restoring the spatial resolution of the feature maps. Let the input feature map be $X \in R^{H' \times W'}$, the convolution kernel be $K \in R^{K_h \times K_w}$, and the output feature map be $Y \in R^{H \times W}$.

$$\begin{cases} y(i,j) = \sum_{m=0}^{K_h-1} \sum_{n=0}^{K_w-1} x(i-m, \ j-n) \cdot \omega(m,n) \\ H = (H'-1) \times S + K_h \\ W = (W'-1) \times S + K_w \end{cases} \tag{5}$$

In this study, the primary role of deconvolution is to perform up-sampling: by applying deconvolution to the Gabor-ASTFT feature maps, we achieve higher resolution and resize the feature maps to an appropriate dimension, facilitating effective feature fusion.

### 3.4. Dilated Residual layers

Dilated Residual layers are a convolutional neural network structure that combines the advantages of residual structures and dilated convolutions. A feature learning model based on this structure is constructed for fault classification.As the depth of neural networks increases, deep models may suffer from the degradation problem. During backpropagation, as the order of chain differentiation increases, gradient vanishing or gradient explosion may occur. These issues prevent the model's representation ability from improving with increasing network depth and may even have the opposite effect. Sergey Ioffe and Christian Szegedy proposed the Batch Normalization (BN) layer to address gradient vanishing and explosion[28]. In Resnet model porposed by Kaiming He, the extensive use of BN layers and residual structures effectively mitigated the model degradation problem[29].Therefore, based on residual structures and BN layers, a residual network is constructed to perform deep feature extraction and fault classification on dual-channel fused features. The residual expression is formulated

as:

$$\vec{x} \rightarrow \vec{y} = F(\vec{x}) + \vec{x} \tag{6}$$

where $F(\vec{x})$ represents the learned features, $\vec{x}$ is the original feature (input from the previous layer), and $\vec{y}$ is the final output.By incorporating skip connections, the output features include both identity mapping and residual mapping, effectively addressing the gradient vanishing and degradation problems.

The inclusion of BN layers alleviates internal covariate shift during training, accelerating the learning process and enhancing model stability. This enables the network to achieve better robustness and representational capacity. The BN layer primarily performs operations such as computing the mean $\mu_B$ and variance $\sigma_B^2$, normalizing data $\hat{x}_i$, and introducing learnable scaling $\gamma$ and shifting $\beta$ parameters.

$$\begin{cases} \mu_{\mathrm{B}} = \dfrac{1}{\mathrm{m}} \sum_{i=1}^{m} x_i \\[2mm] \sigma_B^2 = \dfrac{1}{\mathrm{m}} \sum_{i=1}^{m} (x_i - \mu_{\mathrm{B}})^2 \\[2mm] \hat{x}_i = \dfrac{x_i - \mu_{\mathrm{B}}}{\sqrt{\sigma_B^2 + \varepsilon}} \\[2mm] y_i = \gamma x_i + \beta \end{cases} \tag{7}$$

In the residual layers, after convolutional processing, the obtained feature maps undergo a linear mapping. To enhance the model's expressive capability, an activation function is introduced to transform the linear data into nonlinear data, enabling the composite function model to better fit the data and handle more complex problems. Therefore, in the intermediate hidden layers, the commonly used ReLU function is employed as the activation function:

$$ReLU(x) = max(0, x) \tag{8}$$

In the final layer, the softmax function is utilized as the classification function:

$$\sigma\left(\vec{z}\right)_i = \frac{e^{z_i}}{\sum_{j=1}^{K} e^{z_j}} \tag{9}$$

The feature maps generated after convolution and activation are often large, which increases the model's training time and may lead to overfitting. To address this issue, pooling layers are incorporated to compress the extracted feature data, thereby reducing the computational complexity of the network and decreasing training time. Two types of pooling layers are introduced: after the input data undergoes dual-channel feature fusion, the feature maps are fed into a max pooling layer, which partitions the feature maps into multiple rectangular regions while preserving spatial structure and extracting the local maximum values to construct new feature maps. This process retains the most representative features, eliminates redundant data, reduces the number of parameters, and optimizes the network model.

$$y^l[i] = \max_{(i-1)k+l \leq n \leq ik} \{y^{l-1}[n]\} \tag{10}$$

where y l [i] represents the output of i-th pooling region, and k is the length of the kernel of pooling layer.

Another approach is Global Average Pooling (GAP), which replaces the fully connectedlayer in the final layer of the model. This reduces redundant parameters, decreases computational complexity, and enhances the model's generalization capability.

$$GAP_c = \frac{1}{H \times W} \sum_{i=1}^{H} \sum_{j=1}^{W} X_c(i,j) \tag{11}$$

where $GAP_c$ represents the GAP output value for the c channel, $X_c(i,j)$ denotes the input, and $H \times W$ is the size of the feature map.

In the ResNet architecture, standard convolution is utilized as the primary convolutional operation. In strided convolutional layers with a stride of 2, the output feature map size is reduced to half of the input feature map size. As the network depth increases, feature maps undergo multiple convolutional operations, ultimately resulting in a small feature map (1×1) at the network's final stage. Although ResNet preserves most feature information by increasing the number of convolutional channels, the reduced resolution of the final feature map leads to partial feature loss, which, to some extent, degrades classification accuracy.

To improve the resolution of the final feature map, a common approach is to reduce the number of downsampling layers, specifically by decreasing the number of strided convolutional layers. However, this approach reduces the receptive field. To address this issue, dilated convolution is introduced as a replacement for certain strided convolutional layers, increase the receptive field of the higher layers, compensating for the reduction in receptive field induced by removing subsampling[30].

By replacing strided convolution with dilated convolution, the model significantly enlarges its receptive field without increasing the number of parameters, enhances feature extraction capability, mitigates the resolution degradation caused by strided convolution, and improves the model's ability to capture fine details, thereby enhancing classification performance. Taking an input feature map of 224 × 224 as an example, without pooling or downsampling, the feature map changes after four convolutional layers are shown in Figure 3. As observed, compared to strided convolution, dilated convolution retains a higher level of spatial resolution in the feature maps.

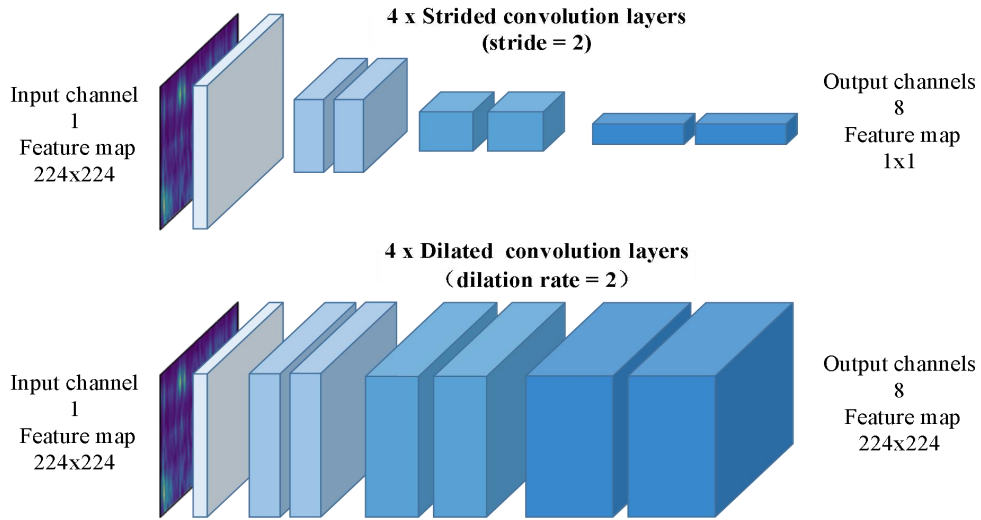

**Figure 3.** Diagram of Feature Map Size Changes Under Different Convolutions

## 4. The Proposed Method：

Based on the aforementioned theory, On the basis of the theories mentioned above, an integrated intelligent method of fault diagnosis for gears using acceleration signals was proposed, which was based on Gabor-ASTFT and DTCWT algorithms,Dilated Residual structure and feature fusion layer.The main process of the proposed method is illustrated in Figure 4,and the key steps are as follows:

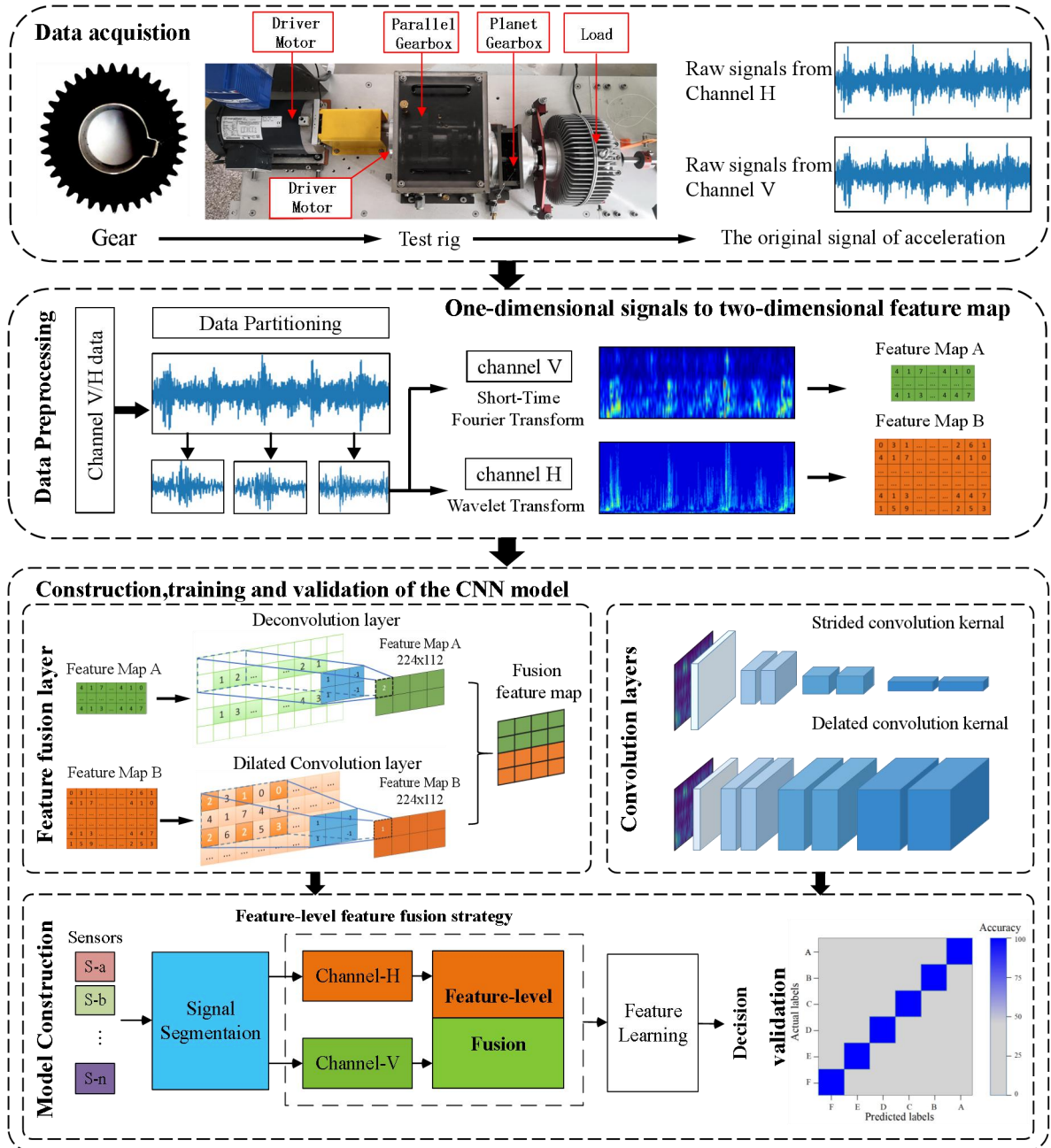

**Figure 4.** The proposed fault diagnosis method of gear.

Step 1: Collect acceleration signals from both the horizontal and vertical channels at the first gear of the gearbox, where the faulty gear is located, on the test bench. By replacing gears with different fault types, multi-channel fault data corresponding to various fault types are obtained.

Step 2: Acquire sufficient raw signal samples. Based on the motor speed, sampling frequency, and the transmission ratio of the gear system, segment the collected raw signals into appropriately sized signal fragments. The segmented signals are then normalized and processed using the Gabor-ASTFT and DTCWT. The resulting two-dimensional feature signals are used to construct the training, validation, and test datasets.

Step 3: Referencing the Dilated Residual Networks, a convolutional neural network

(CNN) model incorporating a feature fusion layer is established. In this model, dilated convolution and transposed convolution are employed to resample the signal features from different channels to a suitable size before feeding them into the fusion layer for feature integration.

Step 4: Utilize the dilated residual structure to perform deep feature extraction. Finally, a Softmax classifier is applied for fault classification. The classification results and model performance are compared and analyzed.

Step 5: Train the constructed CNN model using different datasets and evaluate the model's accuracy and generalization performance across various datasets.

## 5. Experimental Validations

### 5.1 Data Information

In this study,two datasets from the open-source WT-planetary gearbox in Beijing University Of Technology(BJUT) and the wind turbine drivetrain diagnostics simulator (WTDS) in the author's laboratory are adopted to verify the proposed fault diagnosis method for rolling bearing.

The data were collected from the wind turbine drive train test rig, with the acceleration signal of the motor-driven gear sampled at 48 kHz under a motor speed of 30 Hz（1800rpm）. The vibration data were obtained using a Sinocera CA-YD-1181 accelerometer, and an encoder was employed to capture the speed pulses. The collected signals were segmented into fixed-length fragments based on the fault frequency of sun gear (25/8)，with the number of sampling points set to 1536, calculated according to the gear's transmission ratio and motor speed.

The dataset includes five health conditions of the sun gear. For the classification task, these five fault categories were selected, and the dataset consists of 5,000 samples used for model training, validation, and testing. The details of the dataset are shown in Table 1.

**Table 1.** Details of the experimental dataset collected from the BJUT .

| Label | Fault information | Length of Sample | Count of Samples | Training Set | Validation Set | Testing Set |
|-------|-------------------|------------------|------------------|--------------|----------------|-------------|
| H | Healthy condition | | | | | |
| G | gear with a broken tooth | | | | | |
| W | wear gear | 1536 | 5000 | 3000 | 1000 | 1000 |
| C | crack occurs in the root | | | | | |
| M | missing one tooth | | | | | |

In the other dataset, the acceleration signals collected from the WTDS test rig have a sampling frequency of 48,000 Hz, with the motor drive speed set to 25Hz(1500 rpm) and a load of approximately 45 Nm. Additionally, the dataset contains two channels of acceleration signals corresponding to the horizontal and vertical sensor installations. There is also one channel of tachometer pulse signals in the dataset. Similar to the first dataset, the raw signals are segmented into fixed-length fragments.The sampling duration is 0.1s ((1n*90/36)/25hz) when the fault gear rotates for one turn, with each fragment having a length of

2048(20480hz*0.1s) sampling points.

The human-made faults are located respectively on the different place of gears. Therefore, a total of four types of signals, with one normal and three faults, are contained in the dataset. The details can be seen in Table 2.

**Table 2.** Details of the experimental dataset collected from the author's laboratory.

| Label | Fault information | Length of Sample | Count of Samples | Training Set | Validation Set | Testing Set |
|-------|-------------------|------------------|------------------|--------------|----------------|-------------|
| H | Healthy condition | | | | | |
| W | wear gear | | | | | |
| M1 | missing 2/3 tooth | 2048 | 4800 | 2880 | 960 | 960 |
| M2 | missing tooth | | | | | |
| C | crack occurs in the root | | | | | |
| E | eccentric gear | | | | | |

The time domain waveforms and envelope spectrums of the signals in WTDS's dataset are shown in Figure 5.By the figure,the weak features of the fault signal are partially manifested in the original signal and the differences among the different faults exist but cannot be diagnosed accurately. Therefore, more features in the raw signals need to be extracted for the fault diagnosis.

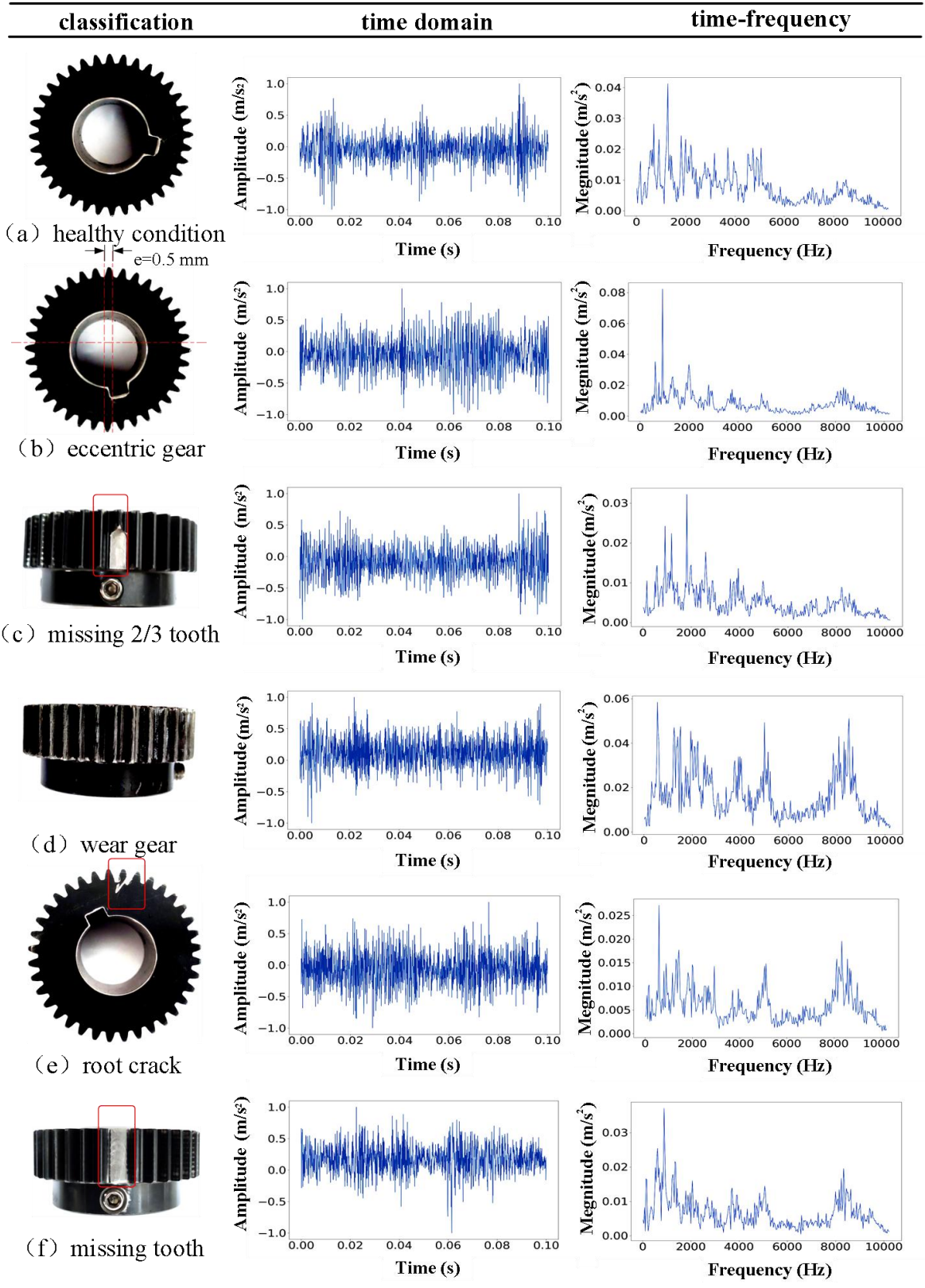

**Figure 5.** Gears and signals

## 5.2 Preprocessing of Signals

When a component of rotating machinery fails, it may also affect other working components. If a gear fails, cyclical impact modulation may appear in the vibration signal at specific fault frequencies. Typically, the acceleration signals collected from the gearbox base are non-stationary and contain significant noise. Additionally, the presence of many complex modulation components in the signal makes the fault analysis process more challenging, making it difficult to accurately classify the types of faults occurring in rotating machinery.

In this paper, the proposed Gabor ASTFT and DTCWT algorithms are used to extract more features from the noisy raw signals, thereby transforming the one-dimensional time-domain signals into two-dimensional time-frequency features for model classification. The results of Gabor ASTFT and DTCWT in Figures 6-7 show that after transformation, different types of fault features become more pronounced. For example, when tooth breakage occurs, the gear mesh frequency (GMF) and its harmonics show sudden amplitude changes, and the periodic impact caused by missing teeth leads to sidebands near the GMF. In the case of gear eccentricity, low-frequency components are enhanced. When gear surface wear occurs, the energy in the high-frequency region increases, but no obvious local impact features are observed, and the amplitude changes remain relatively smooth. In the normal state, the gear signal has a uniform energy distribution with clear periodicity. However, it is still challenging to accurately determine the fault type based solely on the time-frequency map. Further extraction of fault features is needed.

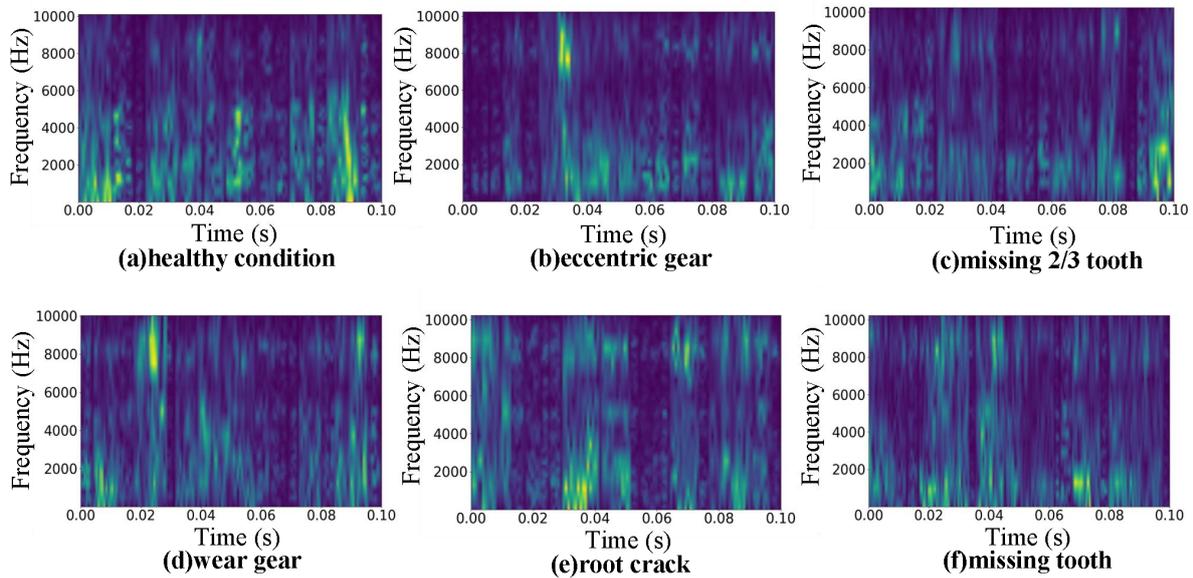

**Figure 6.** Gabor-ASTFT time-frequency diagram

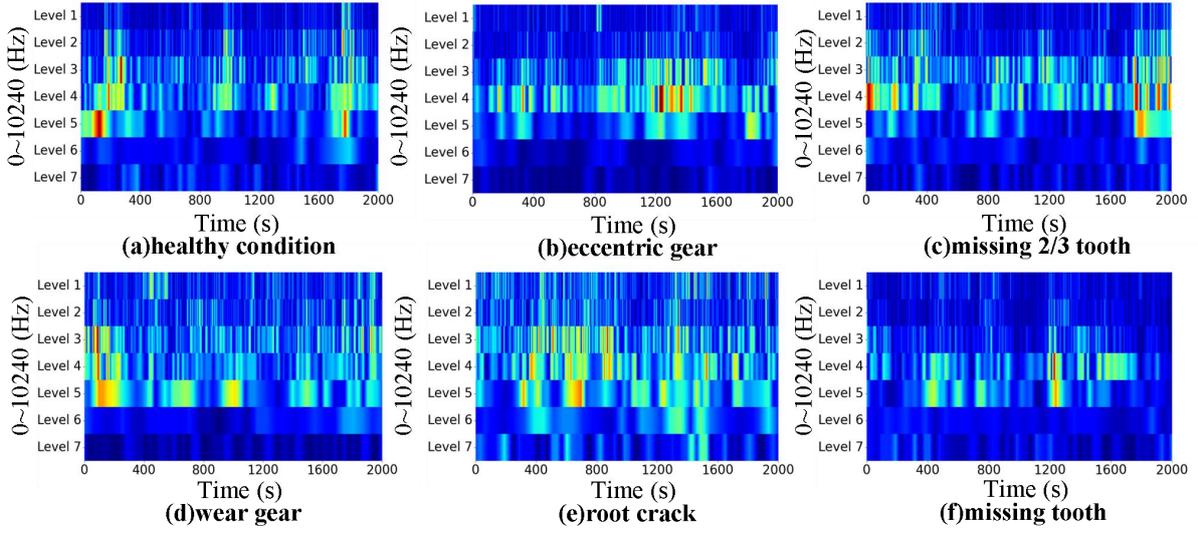

**Figure 7.** DTCWT time-frequency diagram

## 5.2 Case Study I

The training parameters are the same for both datasets, using the Adam optimizer, batch size of 32, 20 iterations, and an initial loss value of 0.0001.

In the first section, the BJUT dataset is used as the input data for the proposed CNN model. As shown in Table 1, the gear fault conditions include five types, with 1000 samples for each condition. Each sample consists of 1536 acceleration values. The dataset is divided into training, validation, and test sets in a ratio of 6:2:2. Additionally, the maximum number of training iterations is set to 25.

During training, the one-dimensional acceleration time-domain signals from both the horizontal and vertical channels are simultaneously fed into the model. These signals undergo Gabor-ASTFT transformation and are processed through an upsampling transposed convolutional channel, while the DTCWT transformation is applied through a downsampling dilated convolutional channel. The extracted features are then fused and input into the residual structure for model training.

Based on the trained model, the classification performance is evaluated using the test set. After multiple classification experiments, the proposed CNN model achieves an average classification accuracy of 100%, indicating that the model has reached optimal training performance and demonstrates excellent fault classification capability.

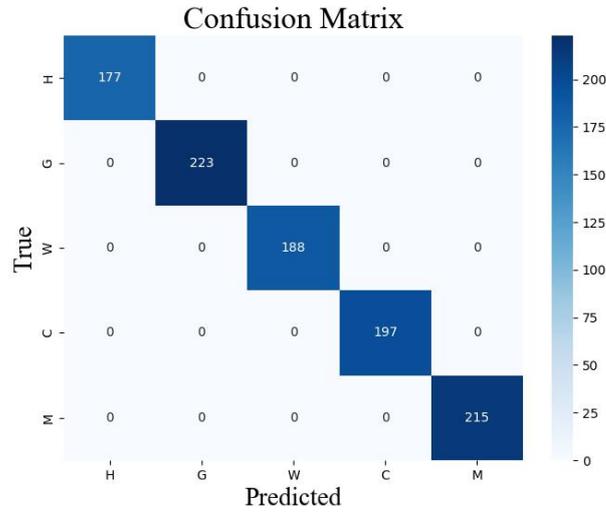

**Figure 8.** Classification result of BJUT's dataset.

To verify the effectiveness of the feature fusion strategy and the Gabor-ASTFT and DTCWT transformation algorithm, the model was adjusted to five different types: single-channel input of raw signals, single-channel input of extracted feature signals, and dual-channel feature fusion. Multiple comparative experiments were conducted, recording the loss values and accuracy values during the training process. The corresponding loss function curves and accuracy curves are shown in Figure 9-10.The results indicate that the classification accuracy of the model is higher when using Gabor-ASTFT and DTCWT transformed feature signals as input compared to directly inputting raw signals. This demonstrates that preprocessing vibration signals with Gabor-ASTFT and DTCWT contributes to improving the training accuracy and convergence speed of the model. Additionally, it is observed that the classification accuracy of the model further improves when the extracted feature signals undergo feature fusion, confirming that the feature fusion strategy enhances the model's fault diagnosis capability.

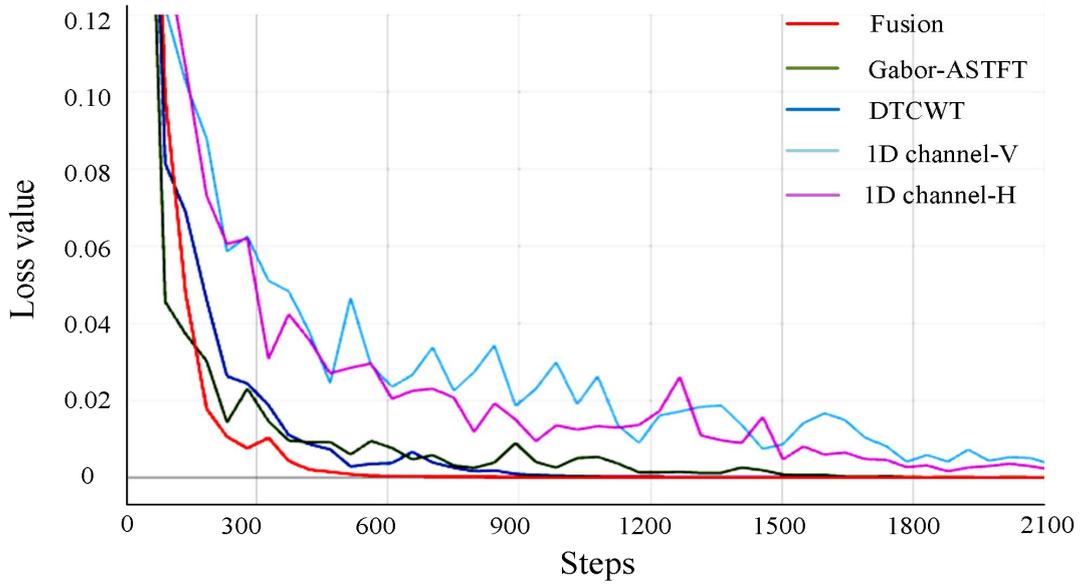

**Figure 9.** Loss validation curves of the proposed 1D-CNN model based on BJUT's dataset.

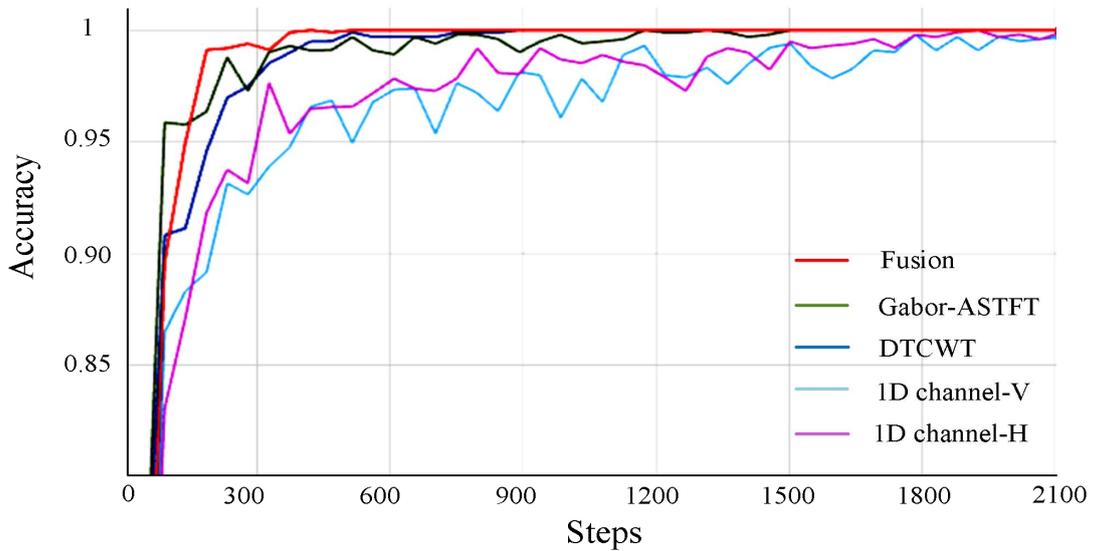

**Figure 10.** Accuracy curves of the proposed 1D-CNN model based on BJUT's dataset.

The accuracy and training time of the model on the test set are shown in Table 3. The classification accuracy demonstrates the effectiveness of the signal preprocessing process and the feature fusion strategy. The results indicate that data preprocessing and feature fusion strategies improve the classification ability of the model.However, since the feature fusion model utilizes dual-channel input, it requires more training data and has a larger number of model parameters. As a result, the training time of the feature fusion model is significantly increased.

**Table 3.** Details of the experimental dataset collected from BJUT .

| Model Names | Accuracy | Time(s) |
|---|---|---|
| Proposed model with processed feature fusion | 100% | 140.16 |
| Resnet-18 with 2D signals(Gabor-ASTFT) from channel V | 99.6% | 59.29 |
| Resnet-18 with 2D signals(DTCWT) from channel H | 98.4% | 74.76 |
| Resnet-18 with raw signals from channel V | 99.15% | 45.93 |
| Resnet-18 with raw signals from channel H | 97.15% | 47.24 |

*5.2 Case Study II*

In the second section, the WTDS dataset is used as the input data for the proposed model. As shown in Table 2, the gear fault states consist of six types, with one additional fault state compared to the BJUT dataset. Each fault type contains 800 samples, with a length of 2048 acceleration values. The dataset is also divided into training, validation, and test sets in a 6:2:2 ratio. Additionally, the maximum number of training iterations is set to 25.

The training process follows the same methodology as that used for the BJUT dataset and yields similar experimental results. The confusion matrix of the model's classification performance on the test set is shown in Figure 11.After multiple classification experiments, indicating that the proposed model also performs well on the WTDS dataset, achieving a classification accuracy of 100%.

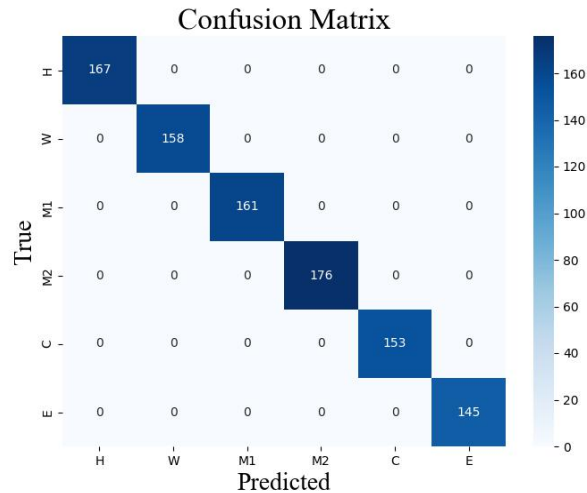

**Figure 11.** Classification result of WTDS's dataset.

Similarly, to verify the effectiveness of the feature fusion strategy and the Gabor-ASTFT and DTCWT transformation algorithm, the model was adjusted into five different configurations, following the same experimental setup as on the BJUT dataset. Multiple comparative experiments were conducted, and the loss values and accuracy values during the training process were recorded. The corresponding loss function curves and accuracy curves are shown in Figure 12-13.The results indicate that, compared to directly inputting the raw signal, using Gabor-ASTFT-transformed and DTCWT-transformed feature signals as input leads to higher classification accuracy. This demonstrates that preprocessing vibration signals with Gabor-ASTFT and DTCWT enhances both the training accuracy and the convergence speed of the model. By employing a fusion strategy, the model exhibits a faster learning process and a more stable convergence curve.

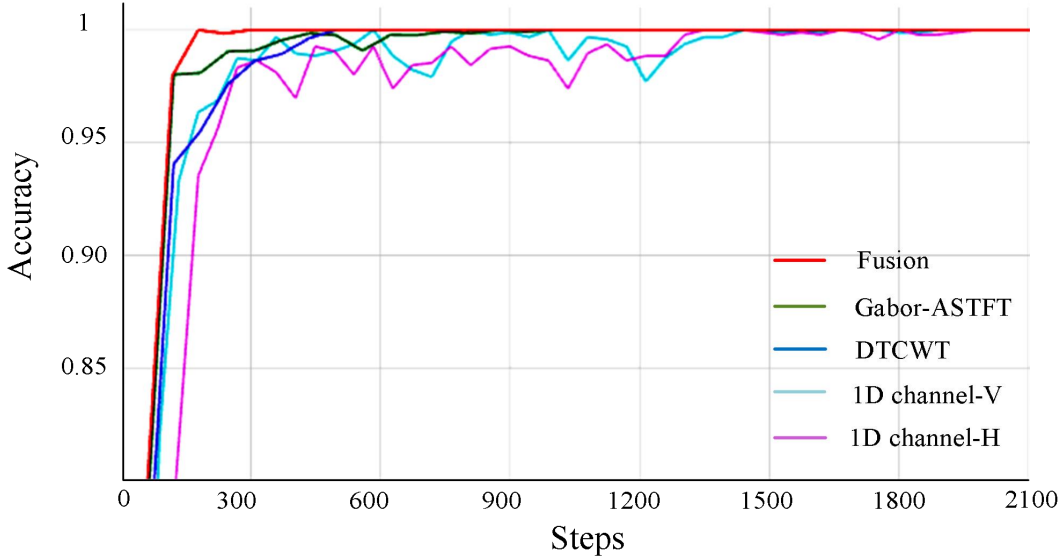

**Figure 12.** Loss validation curves of the proposed 1D-CNN model based on WTDS's dataset.

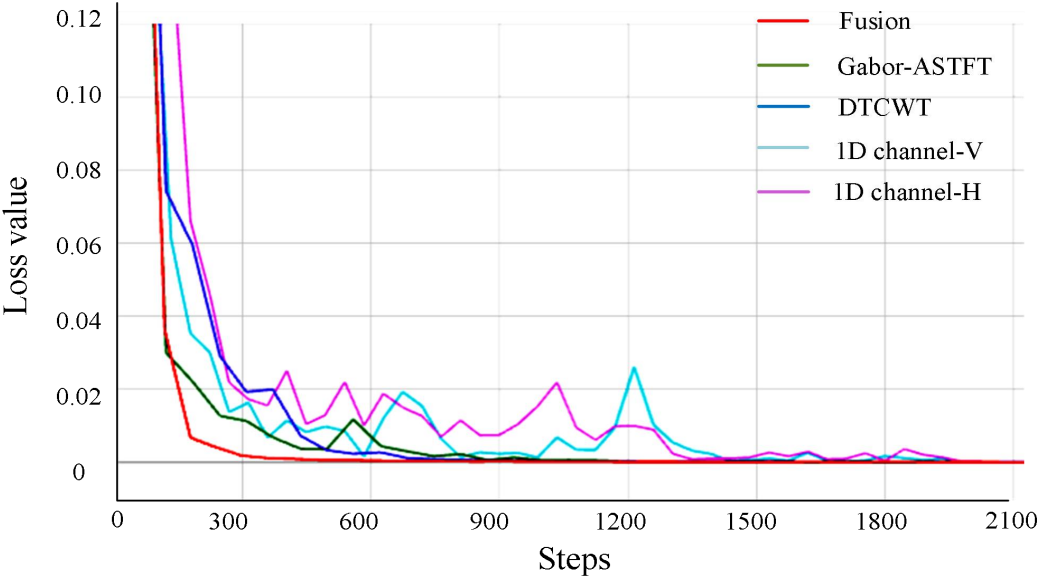

**Figure 13.** Accuracy curves of the proposed 1D-CNN model based on WTDS's dataset.

However, as shown in Table 4, after performing feature fusion on the extracted feature signals, the classification accuracy of the model does not further improve, as it has already reached 100%. Nonetheless, a comparison of the convergence curves during model training reveals that the loss curve of the fused features declines more rapidly, while the accuracy curve rises more steadily. This suggests that the feature fusion strategy still contributes to the improvement of training performance to some extent, even though its effect is not directly reflected in the classification accuracy on the test set.

**Table 4.** Details of the experimental dataset collected from author's laboratory .

| Model Names | Accuracy | Time(s) |
|---|---|---|
| Proposed model with processed feature fusion | 100% | 131.88 |
| Resnet-18 with 2D signals(Gabor-ASTFT) from channel V | 100% | 80.94 |
| Resnet-18 with 2D signals(DTCWT) from channel H | 100% | 99.60 |
| Resnet-18 with raw signals from channel V | 99.8% | 54.93 |
| Resnet-18 with raw signals from channel H | 99.35% | 52.75 |

## 6. Conclusions

In this study, an integrated intelligent method of fault diagnosis for gears using acceleration signals was proposed, which was based on Gabor-ASTFT and DTCWT algorithms, Dilated Residual structure and feature fusion layer.

Within the proposed method, acceleration signals from both the horizontal and vertical channels of a planetary gearbox test rig are collected. Gabor-ASTFT and DTCWT are applied to extract features from different channels, converting the 1D raw vibration acceleration signal into a 2D time-frequency representation. Deconvolution and dilated convolution operations are then used to upsample and downsample the 2D feature maps to adjust their sizes. A dual-channel 2D-CNN model with a feature fusion layer is constructed to integrate features and perform fault classification. To evaluate the proposed approach, comparative experiments are conducted on different datasets, leading to the following conclusions:

The Gabor-ASTFT and DTCWT algorithms effectively extract fault-related features from raw gear acceleration signals, significantly improving the classification accuracy of the model.

The application of dilated convolution and deconvolution enhances the model's generalization capability by enabling upsampling and downsampling operations. This allows feature maps of varying sizes to be adjusted while preserving essential signal characteristics, facilitating seamless feature fusion.

The feature fusion layer expands the model's capability for extracting fault-related information from gear signals by enabling multi-channel analysis of the same fault signal. This enhances fault feature extraction and further improves classification accuracy.

Experimental results based on different datasets demonstrate the effectiveness, accuracy, and stability of the proposed method, providing an end-to-end solution for gear fault diagnosis.

Future research will explore the broader applicability of this approach, including its extension to variable operating conditions and the diagnosis of compound faults in gears and other rotating machinery.